%% file: main.tex
\documentclass[runningheads]{llncs}

\usepackage[T1]{fontenc}
\usepackage{graphicx}
\usepackage{color}
\usepackage{makecell}
\usepackage{arydshln}
\usepackage{listings}
\usepackage{tcolorbox}
\usepackage{cleveref}
\usepackage{multirow}

\newtcolorbox[auto counter, number within=section]{promptbox}[2][]{%
    colback=gray!5, colframe=black, fonttitle=\bfseries,
    title=Prompt~\thetcbcounter: #2, #1
}

\crefformat{figure}{#2#1#3} 
\crefrangeformat{figure}{#3#1#4--#5#2#6} 
\crefmultiformat{figure}{#2#1#3}{ and #2#1#3}{, #2#1#3}{, and #2#1#3}  

\begin{document}

\title{Explaining Low Perception Model Competency with High-Competency Counterfactuals}

\titlerunning{Explaining Competency with Counterfactuals}

\author{Sara Pohland\orcidID{0000-0003-2746-6372} \and
Claire Tomlin\orcidID{0000-0003-3192-3185}}

\authorrunning{S. Pohland \& C. Tomlin}

\institute{University of California, Berkeley, CA 94720, USA\\
\email{\{spohland,tomlin\}@berkeley.edu}}

\maketitle

\begin{abstract}
There exist many methods to explain how an image classification model generates its decision, but very little work has explored methods to explain why a classifier might lack confidence in its prediction. As there are various reasons the classifier might lose confidence, it would be valuable for this model to not only indicate its level of uncertainty but also explain why it is uncertain. Counterfactual images have been used to visualize changes that could be made to an image to generate a different classification decision. In this work, we explore the use of counterfactuals to offer an explanation for low model competency--a generalized form of predictive uncertainty that measures confidence. Toward this end, we develop five novel methods to generate high-competency counterfactual images, namely Image Gradient Descent (IGD), Feature Gradient Descent (FGD), Autoencoder Reconstruction (Reco), Latent Gradient Descent (LGD), and Latent Nearest Neighbors (LNN). We evaluate these methods across two unique datasets containing images with six known causes for low model competency and find Reco, LGD, and LNN to be the most promising methods for counterfactual generation. We further evaluate how these three methods can be utilized by pre-trained Multimodal Large Language Models (MLLMs) to generate language explanations for low model competency. We find that the inclusion of a counterfactual image in the language model query greatly increases the ability of the model to generate an accurate explanation for the cause of low model competency, thus demonstrating the utility of counterfactual images
in explaining low perception model competency \footnote{The code for reproducing our methods and results is available on GitHub: https://github.com/sarapohland/competency-counterfactuals.}.

\end{abstract}

\input{secs/intro}
\input{secs/background}
\input{secs/generation}
\input{secs/explanation}
\input{secs/conclusion}
\input{secs/future}
\bibliographystyle{splncs04}
\bibliography{citations}

\input{secs/appendix}

\end{document}

%% file: secs/intro.tex
\section{Introduction} \label{sec:intro}

Convolutional neural networks (CNNs) have shown impressive performance across a range of image classification tasks, but their black-box nature limits their applicability to real-world systems. Without a thorough understanding of these models and their failure modes, one cannot confidently employ such models for critical decision-making tasks. Within the field of explainable artificial intelligence (xAI), there is extensive work on explaining CNN classification decisions to better understand how models generate their output predictions. However, there has been very limited work on explaining model competency to better understand why a model lacks confidence in its prediction.

Previous work has explored the use of saliency mapping methods to offer explanations for model confidence by identifying particular image regions for which the trained classification model is unfamiliar \cite{xai24}. This is a useful approach when anomalous regions cause the reduction in model competency. However, there are many other non-spatial factors that could lead to a reduction in model confidence, such as changes in image properties like brightness, contrast, or saturation, as well as holistic image corruption like noise or pixelation. We need other methods to offer explanations for low model competency in these cases.

We explore the use of counterfactual images--images associated with high levels of model competency that are similar to the original low-competency image. We develop and compare five approaches for generating counterfactual examples across two distinct datasets with various causes of low model competency. We then evaluate the ability of Multimodal Large Language Models (MLLMs) to generate interpretable explanations for low competency with the aid of these counterfactuals. To our knowledge, this is the first work that uses counterfactual images as an explanatory tool for low model confidence.

%% file: secs/background.tex
\section{Background \& Related Work} \label{sec:background}

In this work, we offer explanations for why an image classification model lacks confidence in its prediction for a given image. There are many methods to quantify model confidence, but we focus on a particular measure of perception model competency, as described in Section \ref{subsec:quantifying}. In Section \ref{subsec:explainable}, we explore many methods employed to explain the predictions of image classifiers and consider their ability to offer explanations for low model competency. Finally, in Section \ref{subsec:language}, we explore the use of language models to expand on these explanations.

\subsection{Quantifying Model Confidence} \label{subsec:quantifying}

CNNs for image classification usually output softmax scores, which can be interpreted as the probability that an image belongs to each of the training classes. The maximum softmax probability (MSP) can serve as a measure of the confidence of the vision model for a given image, but this probability tends to be very close to one \cite{guo} and is particularly unreliable for data outside of the original training distribution \cite{ovadia}. This has motivated many other approaches to quantify model uncertainty, typically through the use of Bayesian Neural Networks (BNNs) \cite{neal-1992,neal_bayesian_1996}, Monte Carlo (MC) dropout \cite{dropout}, or ensembles of models \cite{lakshminarayanan_simple_2017}. These methods capture many aspects of uncertainty, but tend not to capture distributional uncertainty resulting from mismatched training and test distributions \cite{quinonero-candela_dataset_2009}. This has led to methods that specifically seek to detect inputs that are out-of-distribution (OOD), through classification-based \cite{liu_energy-based_2020,liang_enhancing_2020,openmax,dice}, density-based \cite{zong2018deep,rezende_variational_2016,kingma_glow_2018,ren-2019}, distance-based \cite{kl_matching,lee-2018,sun_out--distribution_2022}, or reconstruction-based \cite{xia-2015,gong_memorizing_2019,Sabokrou} approaches. These methods better address distributional uncertainty, but generally rely on thresholds to generate a binary decision, rather than capturing a holistic measure of uncertainty.

We are interested in \textit{perception model competency}--a generalized form of predictive uncertainty that combines various aspects of uncertainty into a single probabilistic score \cite{alice}. To estimate model competency, we employ the PaRCE score \cite{parce}, which computes the product of the MSP and the probability that the image is in-distribution (ID). To estimate the ID probability, PaRCE uses a function of the reconstruction loss of an autoencoder trained to reconstruct the training images. The scores are calibrated via an ID holdout set such that the PaRCE score directly reflects the prediction accuracy of the perception model.

\subsection{Explainable Image Classification} \label{subsec:explainable}

Explainable image classification is a rich field that seeks to offer explanations for why a model makes the decisions that it does \cite{rauker_toward_2023,ali_explainable_2023}. While there are many methods to enhance the understanding of image classifiers, they generally do not deal with \textit{model competency}, and thus cannot offer explanations for how confident a model is in its prediction. We previously explored saliency mapping methods to explain a model's lack of competency, by identifying and displaying key image regions that contribute to the observed low model competency \cite{xai24}. However, while this work offered useful explanations for images with regional features that were unfamiliar to the perception model, there are many other image properties that do not exist at the regional level but may contribute to low competency. To address this limitation, we explore counterfactual methods. 

Rather than seeking to explain why a certain prediction was made, counterfactual methods analyze changes that could be made to the input to obtain a different prediction \cite{counterfactual_review}. Many methods for counterfactual explanations of image classifiers involve pixel-level edits, pinpointing regions for minimal change to achieve the desired class. These approaches frequently use generative models, such as autoencoders, generative adversarial networks (GANs), or diffusion models, to synthesize counterfactual images \cite{explain_gan,generative_liu,Singla2020Explanation,ganterfactual,steex,dime}. There are also a number of optimization-based methods that treat counterfactual generation as a constrained optimization problem in the pixel space \cite{pertinent_negatives,cev_goyal,face,sedc,heads_or_tails}. Similarly, one could design an optimization problem in some latent space with the goal of finding minimal perturbations in the latent representation of an image to effect a change in the classification decision \cite{revise,latent_cf,prototypes,posthoc_guyomard,disc}. Other methods that perform latent space manipulation focus on leveraging the learned semantic structure for interpretable counterfactual generation \cite{dive,c3lt,dcfg,causal_dash}. A diverging line of work uses conceptual counterfactuals, emphasizing human-interpretable semantics. These approaches guide concept-level edits, identifying minimal semantic features that need modification to change a classification \cite{cace,cocox,attribute_based,cce,dissect}. 


While all these methods help explain image classification decisions, none offer explanations for model confidence and all would need to be adapted to varying degrees for this purpose. We explore five novel counterfactual methods that seek to explain why a perception model is not confident for a given image, focusing only on methods that do not require training with low-competency examples.

\subsection{Language Models for Anomaly Explanation} \label{subsec:language}



In our effort to generate counterfactual explanations for low model competency, we explore the use of MLLMs. Although much work has explored the use of visual language models (VLMs) for OOD and anomaly detection \cite{vlm_survey}, often using CLIP \cite{clip} to detect samples that do not belong to any ID class \cite{vlm_ood_ming,vlm_ood_miyai} or to distinguish between normal and abnormal samples \cite{vlm_anomaly_jeong,vlm_anomaly_zhou}, far less work has considered the use of LLMs to provide explanations for anomaly and OOD detection outcomes \cite{llm_ood}. Within the area of LLMs for explanation generation, most work has focused on video anomaly detection (VAD)--the task of identifying unusual or unexpected events in video streams \cite{vad_lv,vad_zhang,vad_zanella,vad_yang}--or time series anomaly detection (TSAD)--the task of identifying unusual patterns or behaviors in time-ordered data points \cite{tsad_zhuang}. We are interested in the use of LLMs to offer explanations of low model competency for individual images, which has yet to be explored. Unlike VAD and TSAD, which analyze temporal changes to detect anomalies, our focus is on identifying spatial features that contribute to uncertainty in a single image.

%% file: secs/generation.tex
\section{Generating Counterfactual Images} \label{sec:generation}

In this work, we explore methods to generate high-competency counterfactual images for low-competency samples and offer explanations for low model competency using MLLMs. In this section, we focus on generating counterfactuals.

\subsection{Counterfactual Generation Methods} \label{subsec:gen-methods}

We develop and compare five distinct methods for generating a high-competency counterfactual image that is qualitatively similar to the original low-competency image. Let $X$ be the original image with a competency score, $\hat{\rho}(X)$, which is below some competency threshold. We hope to generate a counterfactual image, $X'$, whose competency score, $\hat{\rho}(X')$, is above this competency threshold.

\subsubsection{Image Gradient Descent (IGD)} \label{subsubsec:igd}

The goal of the first method is to gradually modify the input image to increase the estimated competency, while maintaining visual similarity.  More specifically, we seek to minimize the loss function
\begin{equation}
    \mathcal{L}(X') = -\hat{\rho}(X') + \gamma d(X,X'),
    \label{eq:igd}
\end{equation}
where $d(\cdot)$ is a distance function and $\gamma$ is a parameter that trades off between increasing the competency of the counterfactual and maintaining similarity between the counterfactual image and the original. We define distance in terms of the Learned Perceptual Image Patch Similarity (LPIPS) metric, which measures how visually similar the original and counterfactual images are \cite{lpips}.

To obtain a counterfactual image that minimizes Equation \ref{eq:igd}, we initially set $X'$ to be the original image, $X$. We then gradually update the image via gradient descent, stopping once the competency of the counterfactual is above the specified threshold or the maximum allowable iterations have been reached.


\subsubsection{Feature Gradient Descent (FGD)} \label{subsubsec:fgd}

In the second method, rather than seeking to maintain visual similarity between the original and counterfactual image, our goal is to increase the estimated competency, while ensuring that the feature vector used for classification does not change substantially. Let $f(\cdot)$ be the feature extractor, which is used to obtain the feature vector, $f(X)$, provided as input to the final softmax layer of the classification model. Our goal now is to minimize
\begin{equation}
    \mathcal{L}(X') = -\hat{\rho}(X') + \gamma d(f(X),f(X')).
    \label{eq:fgd}
\end{equation}
Again, $d(\cdot)$ is a distance function and $\gamma$ is a tunable parameter. By default, we use the negative cosine similarity to represent the distance between two feature vectors, but other distance metrics can be used as well. 

As with IGD, to obtain an image that minimizes Equation \ref{eq:fgd}, we initially set $X'$ to be the original image. We then gradually update the image via gradient descent, stopping once the competency of the counterfactual is above the specified threshold or the maximum allowable iterations have been reached.

\subsubsection{Autoencoder Reconstruction (Reco)} \label{subsubsec:reco}

Recall from Section \ref{subsec:quantifying} that we consider a competency estimation method that relies on an autoencoder to reconstruct the input image \cite{parce}. Because this reconstruction model outputs images similar to those with which it is familiar, we can treat the reconstructed image as a counterfactual. Let $g(\cdot)$ be the encoder of the reconstruction model and $h(\cdot)$ be the decoder. The counterfactual image is then simply $X'=h(g(X))$.


\subsubsection{Latent Gradient Descent (LGD)} \label{subsubsec:lgd}

Improving upon the previous approach, rather than simply using the reconstructed image, we manipulate the latent representation in the reconstruction model to increase the competency of the prediction, while ensuring that the latent vector does not change substantially. Let $z=g(X)$ be the latent representation of the original image and $z'=g(X')$ be the latent representation of the counterfactual image. In this approach, we seek to find the latent vector that minimizes the loss function
\begin{equation}
    \mathcal{L}(z') = -\hat{\rho}(h(z')) + \gamma d(z,z').
    \label{eq:lgd}
\end{equation}
Once again, $d(\cdot)$ is a distance function and $\gamma$ is a tunable parameter. By default, we use the negative cosine similarity to represent the distance between two latent vectors, but other distance metrics can be used as well. 

To obtain a latent vector that minimizes Equation \ref{eq:lgd}, we initially set $z'$ to be the original latent representation, $z$. We then gradually update the latent vector via gradient descent, stopping once the competency of the counterfactual is above the specified threshold or once the maximum number of allowable iterations has been reached. We use the decoder of the reconstruction model to generate the counterfactual image from the latent representation: $X'=h(z')$.


\subsubsection{Latent Nearest Neighbors (LNN)} \label{subsubsec:lnn}

Recall from Section \ref{subsec:quantifying} that competency scores are calibrated via an ID holdout set \cite{parce}. In our final method, we first find the latent vector, $z_{NN}$, from the calibration set that is closest to the latent representation of the image of interest. By default, we use the $\ell_1$ norm to find the nearest neighbor, but other distance metrics may be used as well. We then use the reconstruction of this latent vector as the counterfactual image: $X'=h(z_{NN})$.


\subsection{Comparison of Counterfactual Images} \label{subsec:gen-eval}

We compare our counterfactual image generation methods both quantitatively and visually across two datasets and a number of performance metrics.

\subsubsection{Datasets}

We conduct analysis across two unique datasets. The first dataset is obtained from a simulated lunar environment. The classifier trained on this dataset learns to distinguish between different regions in the environment, such as bumpy terrain, smooth terrain, regions inside a crater, etc. The second dataset contains speed limit signs in Germany \cite{gtsdb}. The classifier learns to distinguish between seven common speed limit signs, ranging from 30 to 120 km/hr. 

While competency tends to be high for both of these datasets, we identify six key causes of low model competency: spatial, brightness, contrast, saturation, noise, and pixelation \cite{parce}. For each dataset, we generate 600 low-competency example images, for which the lack of competency can be attributed to one of these six factors (with 100 images per factor). For the lunar dataset, images with spatial anomalies contain astronauts or human-made structures that were not present in the training set. For the speed limit dataset, spatial anomalies are images of an uncommon speed limit, 20 km/hr, which was not present during training. We generate example images for the other causes of low model competency from high-competency test images by increasing or decreasing the given image property (brightness, contrast, or saturation), adding uniform random noise, or compressing the image to create pixelation. Examples of images with these causes of low model competency are shown in column 1 of Figures \cref{fig:lunar-spatial,fig:lunar-brightness,fig:lunar-contrast,fig:lunar-saturation,fig:lunar-noise,fig:lunar-pixelation} for the lunar dataset and in Figures \cref{fig:speed-spatial,fig:speed-brightness,fig:speed-contrast,fig:speed-saturation,fig:speed-noise,fig:speed-pixelation} for the speed dataset in Appendix \ref{sec:counter_images}.


\subsubsection{Evaluation Metrics} 

Recall from Section \ref{subsec:explainable} that in the field of explainable image classification, counterfactual methods analyze changes that could be made to the input to obtain a different prediction through the generation of counterfactual images. There are five desirable properties of counterfactual images \cite{c3lt}: (I) \textit{Validity} The classification model should correctly assign the counterfactual to the desired class. (II) \textit{Proximity} The counterfactual should remain close to the original in terms of some distance function. (III) \textit{Sparsity} A minimal number of features should be changed in generating the counterfactual. (IV) \textit{Realism} The counterfactual should lie close to the data manifold such that it appears realistic. (V) \textit{Speed} The counterfactual should be generated quickly.

We consider the same properties to be desirable for counterfactuals used to explain why model competency is low for a given image. Rather than defining validity in terms of the classifier's prediction, we say that a counterfactual is valid if the competency estimator assigns it a high competency score. We generate a number of metrics to evaluate our counterfactual generation methods in terms of these properties. (1) \textit{Success rate} To measure validity, we compute the percentage of counterfactuals with high model competency. (2) \textit{Perceptual loss} We evaluate proximity using the LPIPS metric for visual similarity \cite{lpips} described in Section \ref{subsec:gen-methods}. (3) \textit{Feature similarity} We evaluate sparsity in terms of the average cosine similarity between the original and valid counterfactual feature vectors used by the classification model. (4) \textit{Latent similarity} We also evaluate sparsity in terms of the average cosine similarity between the original and valid counterfactual latent representations within the autoencoder of the competency estimator. (5) \textit{Fréchet Inception Distance (FID)} We measure realism first in terms of the FID, which is a metric used to assess the quality of images created by a generative model by comparing the distribution of generated images with the distribution of a set of real images \cite{fid}. The set of real images we use is the set used to calibrate the competency estimator \cite{parce}. (6) \textit{Kernel Inception Distance (KID)} We also assess realism in terms of the KID, which measures the maximum mean discrepancy between features extracted from real and fake images \cite{kid}. (7) \textit{Computation time} Finally, we measure speed in terms of the average time required to compute a counterfactual for a single image.

\subsubsection{Results \& Analysis}

\begin{figure}[h!]
    \centering
    \includegraphics[width=0.95\columnwidth]{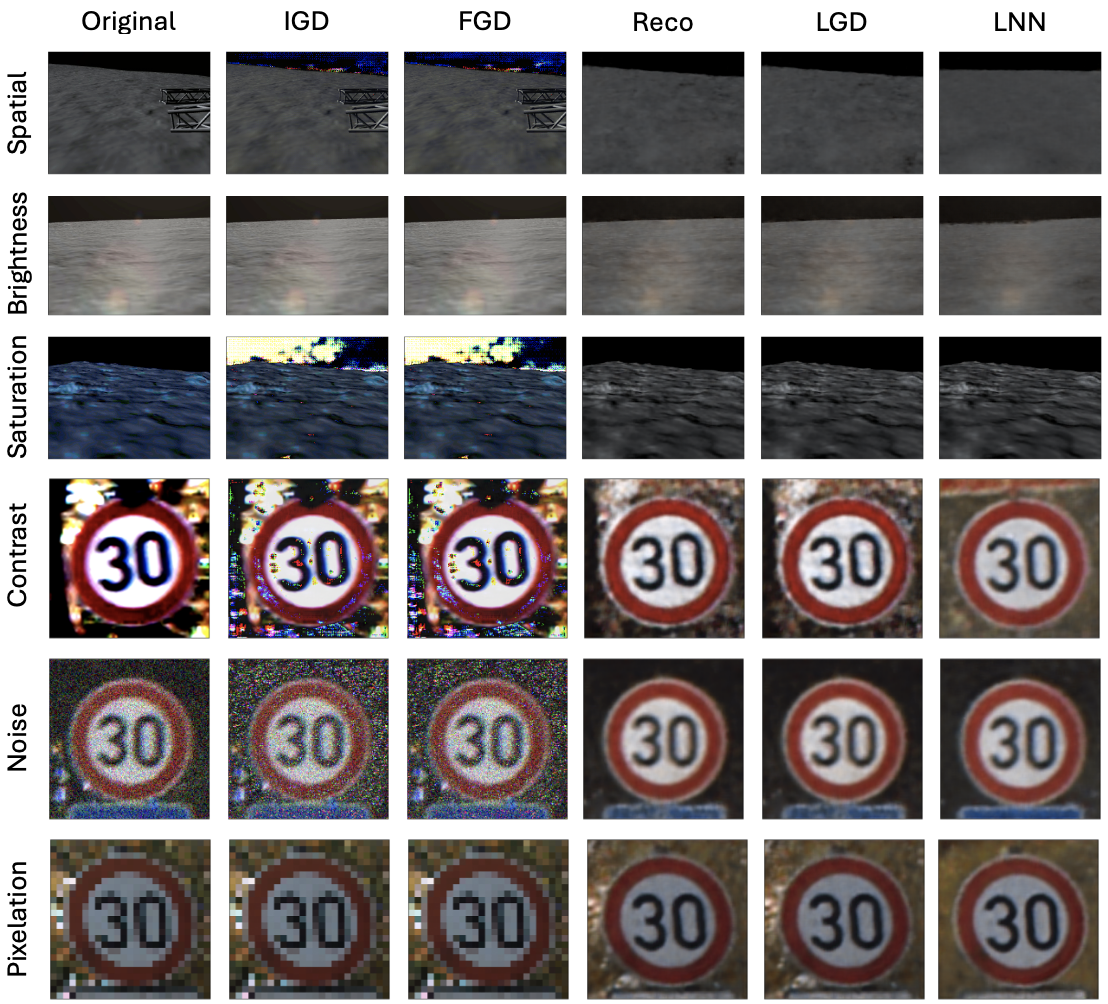}
    \caption{Example counterfactual images generated through different methods (columns) for original images with various causes of low model competency (rows).}
    \label{fig:counterfactual-summary}
\end{figure}

We compare the five counterfactual methods discussed in Section \ref{subsec:gen-methods} for both the lunar dataset (Table \ref{tab:counterfactual-lunar}) and the speed dataset (Table \ref{tab:counterfactual-speed}). For reference, we provide metrics for the original images (Orig) as well. We also visually compare the generated counterfactual images in Figure \ref{fig:counterfactual-summary}. Several additional example images are visualized in Figures \cref{fig:lunar-spatial,fig:lunar-brightness,fig:lunar-contrast,fig:lunar-saturation,fig:lunar-noise,fig:lunar-pixelation,fig:speed-spatial,fig:speed-brightness,fig:speed-contrast,fig:speed-saturation,fig:speed-noise,fig:speed-pixelation} in Appendix \ref{sec:counter_images}.


\vspace{-2mm}
\begin{table}[h!]
    \centering
    \caption{Comparison of counterfactual generation methods for the lunar dataset.}
    \renewcommand{\arraystretch}{1.2}
    \resizebox{\textwidth}{!}{%
    \begin{tabular}{|c|c|c|c|c|c|c|c|}
        \hline
        \makecell{Method} & \makecell{Success \\ Rate $\uparrow$} & \makecell{Perceptual \\ Loss $\downarrow$} & \makecell{Feature \\ Similarity $\uparrow$} & \makecell{Latent \\ Similarity $\uparrow$} & \makecell{\hspace{2mm}FID $\downarrow$\hspace{2mm}} & \makecell{\hspace{2mm}KID $\downarrow$\hspace{2mm}} & \makecell{Computation \\ Time $\downarrow$} \\
        \hline
        Orig & 0.00\% & 0.00 & 1.00 & 1.00 & 10.81 & 21.67 & 0.0002 sec \\
        \hdashline
        IGD & 80.00\% & \textbf{0.02} & 0.98 & 0.97 & 12.39 & 19.38 & 1.1911 sec \\
        FGD & 81.33\% & 0.13 & \textbf{0.99} & 0.97 & 10.97 & 15.82 & 3.1559 sec \\
        Reco & 88.67\% & 0.59 & 0.95 & \textbf{0.98} & 2.63 & \textbf{2.05} & \textbf{0.0053 sec} \\
        LGD & 98.33\% & 0.59 & 0.95 & \textbf{0.98} & 2.61 & 2.06 & 1.0479 sec \\
        LNN & \textbf{99.50\%} & 0.60 & 0.90 & 0.92 & \textbf{2.59} & 2.26 & 0.0069 sec \\
        \hline
    \end{tabular}
    }
    \vspace{-4mm}
    \label{tab:counterfactual-lunar}
\end{table}

\vspace{-2mm}
\begin{table}[h!]
    \centering
    \caption{Comparison of counterfactual generation methods for the speed limit dataset.}
    \renewcommand{\arraystretch}{1.2}
    \resizebox{\textwidth}{!}{%
    \begin{tabular}{|c|c|c|c|c|c|c|c|}
        \hline
        \makecell{Method} & \makecell{Success \\ Rate $\uparrow$} & \makecell{Perceptual \\ Loss $\downarrow$} & \makecell{Feature \\ Similarity $\uparrow$} & \makecell{Latent \\ Similarity $\uparrow$} & \makecell{\hspace{2mm}FID $\downarrow$\hspace{2mm}} & \makecell{\hspace{2mm}KID $\downarrow$\hspace{2mm}} & \makecell{Computation \\ Time $\downarrow$} \\
        \hline
        Orig & 0.00\% & 0.00 & 1.00 & 1.00 & 29.79 & 82.23 & 0.0001 sec \\
        \hdashline
        IGD & 98.33\% & \textbf{0.01} & 0.54 & \textbf{0.99} & 83.64 & 315.99 & 2.8882 sec \\
        FGD & 95.83\% & 0.02 & \textbf{0.81} & \textbf{0.99} & 80.11 & 297.85 & 5.4005 sec \\
        Reco & 74.67\% & 0.49 & 0.59 & 0.88 & 8.65 & 8.23 & \textbf{0.0140 sec} \\
        LGD & \textbf{100.00\%} & 0.47 & 0.56 & 0.88 & \textbf{8.48} & \textbf{8.12} & 4.2700 sec \\
        LNN & 91.33\% & 0.53 & 0.41 & 0.58 & 9.18 & 8.72 & 0.0159 sec \\
        \hline
    \end{tabular}
    }
    \vspace{-4mm}
    \label{tab:counterfactual-speed}
\end{table}

Comparing the five counterfactual generation methods across both datasets, we first observe that LGD most reliably generates high-competency counterfactual images, achieving a 100\% success rate on the speed limit dataset and nearly 100\% success for the lunar dataset. LNN also achieves nearly 100\% on the lunar dataset but its success rate is closer to 90\% for the speed dataset. IGD and FGD tend to perform similarly, generating high-competency counterfactuals for over 95\% of the low-competency images in the speed limit dataset but around 80\% for the lunar dataset. Finally, Reco is close to 90\% successful for lunar but only  around 75\% for speed, indicating that this method is not the most reliable.

Comparing the proximity of counterfactual images to the original images, we notice that IGD performs the best in terms of perceptual loss, followed by FGD. Reco, LGD, and LNN perform similarly with higher perceptual losses. 

We also see similar results for sparsity, for which we consider changes in both the feature vectors and latent representations of the original low-competency images. We find that FGD tends to produce counterfactual images with the most similar feature vectors, while both IGD and FGD produce counterfactuals with very similar latent representations. We also notice that, overall, more elements of the original feature vectors and latent representations are changed with LNN.

However, we see nearly opposite results in terms of realism. We observe that Reco, LGD, and LNN produce counterfactual images that are much more realistic than the original low-competency images, with little difference between these three methods. In contrast, IGD and FGD produce counterfactuals that are similarly unrealistic to the original images or sometimes even more unrealistic. 


The visual comparison of these methods sheds some light on these quantitative results. From rows 1, 3, and 4 of Figure \ref{fig:counterfactual-summary}, we observe that IGD and FGD sometimes produce counterfactual images with unrealistic artifacts. It is also clear that IGD and FGD often produce counterfactuals that are proximal, but this is not necessarily achieved in a positive way. As is observed in rows 2 and 6 of Figure \ref{fig:counterfactual-summary}, the differences between the original and counterfactual images are not always clearly observable, which is not beneficial for an explanatory tool.

Finally, comparing the speed of the five methods, we observe that Reco is the fastest on average, but LNN is similarly fast. IGD and LGD are significantly slower than these two methods, and FGD tends to be the slowest. It is also interesting to note that computation time varies significantly with the dataset.

Returning to our visual comparison (Figure \ref{fig:counterfactual-summary}), we see that Reco, LGD, and LNN often produce counterfactuals that correct the cause of low model competency observed in the original images. In Figure \cref{fig:lunar-spatial}, we see objects were removed from examples with spatial anomalies, and in Figure \cref{fig:speed-spatial}, we observe the digit 2 associated with an unfamiliar class was replaced with a digit associated with a seen class. Similarly, Figures \cref{fig:lunar-brightness,fig:speed-brightness} demonstrate that brightness of overexposed images is corrected in the counterfactuals, contrast for high-contrast images is reduced in Figures \cref{fig:lunar-contrast,fig:speed-contrast}, and saturation is reduced for overly saturated images in Figures \cref{fig:lunar-saturation,fig:speed-saturation}. We also notice that noise was removed from noisy images in Figures \cref{fig:lunar-noise,fig:speed-noise}, and pixelation was corrected in \cref{fig:lunar-pixelation,fig:speed-pixelation}.

In general, the ``best'' method depends on which properties of counterfactual image generation are valued most highly. IGD and FGD are probably not particularly useful because they often produce unrealistic images that generally do not address the true cause of low model competency. However, they would be useful if proximity and similarity are a major concern. If speed is a high priority, one might opt for Reco or LNN over LGD. However, if it is most important to reliably produce high-competency counterfactuals, then LGD should be chosen instead. The appropriate method largely depends on how the counterfactual will be used. In the next section, we consider how these counterfactuals might be used by an MLLM to generate language explanations for low model competency.

%% file: secs/explanation.tex
\section{Explaining Counterfactual Images}  \label{sec:explanation}

In this section, we consider how to obtain language explanations for low model competency using the counterfactual images generated in Section \ref{sec:generation}.

\subsection{Counterfactual Explanation Method} \label{subsec:expl-method}

We focus on explaining potential causes for low model competency with the help of MLLMs. We consider the explanation provided when the model sees only the original image, as well as the explanation provided upon seeing both the original and counterfactual image. Based on our results from Section \ref{subsec:gen-eval}, we use the autoencoder reconstruction (Reco), latent gradient descent (LGD), and latent nearest neighbors (LNN) methods to generate counterfactual images.

\subsubsection{LLaMA Model} 

While a number of MLLMs were considered for the purpose of counterfactual explanation, all explanations are generated using the LLaMA 3.2 model (in the 11B size) \cite{llama} because it is a publicly available model that has demonstrated strong performance in Visual Question Answering (VQA) tasks \cite{vqa_eval}. The LLaMA 3.2 model is a pre-trained and instruction-tuned image reasoning generative model that is optimized for visual recognition, image reasoning, captioning, and answering general questions about an image. This model allows one to set the context in which to interact with the AI model, which typically includes rules, guidelines, or necessary information that help the model respond effectively. It also allows for user prompts, which include the inputs, commands, and questions to the model that could contain an image with text or text only.

\subsubsection{Model Prompts} 

To obtain an explanation from the language model of low model competency for a given image, we first describe the training set, using Prompt \ref{prompt:dataset_lunar} for the lunar dataset and Prompt \ref{prompt:dataset_speed} for the speed limit dataset. We also give a description of the competency estimator using Prompt \ref{prompt:competency}. We then provide instructions about the desired output, using Prompt \ref{prompt:instruct_none} if we are not using a counterfactual image and Prompt \ref{prompt:instruct_counter} otherwise.


\subsection{Comparison of Counterfactual Explanations} \label{subsec:expl-eval}


For each language model explanation, we manually evaluate whether the response correctly describes the true cause of low model competency. We compare the correctness of the explanations that do not use counterfactual images to those aided by the counterfactuals generated by the Reco, LGD, and LNN methods. The accuracies of the explanations across each of the six causes of low model competency, along with the average accuracy, are provided for the lunar dataset in Table \ref{tab:explanations-lunar} and the speed limit dataset in Table \ref{tab:explanations-speed}. Note that we primarily assess the performance of the pre-trained LLaMA model in generating appropriate explanations, but we report the performance for a fine-tuned model as well.

\vspace{-2mm}
\begin{table}[h!]
    \centering
    \caption{Accuracy of competency explanations for the lunar dataset across various true causes of low model competency. Results for the pre-trained model are displayed more prominently, while results for the fine-tuned model are provided in parentheses.}
    \renewcommand{\arraystretch}{1.2}
    \begin{tabular}{|c|c|c|c|c|c|c|c|}
        \hline
        \hspace{1mm}Method\hspace{1mm} & \hspace{1mm}Spatial\hspace{1mm} & \hspace{1mm}Brightness\hspace{1mm} & \hspace{1mm}Contrast\hspace{1mm} & \hspace{1mm}Saturation\hspace{1mm} & \hspace{1mm}Noise\hspace{1mm} & \hspace{1mm}Pixelation\hspace{1mm} & \hspace{1mm}Average\hspace{1mm} \\
        \hline
        None & 8\% & 1\% & 6\% & 1\% & 6\% & \textbf{91\%} & 18.83\% \\
            [-0.75ex]
             & {\tiny (99\%)} & {\tiny (90\%)} & {\tiny (100\%)} & {\tiny (100\%)} & {\tiny (100\%)} & {\tiny (100\%)} & {\tiny (98.17\%)} \\
        \hline
        Reco & \textbf{28\%} & \textbf{10\%} & \textbf{13\%} & 7\% & 73\% & 77\% & 34.67\% \\
            [-0.75ex]
             & {\tiny (95\%)} & {\tiny (83\%)} & {\tiny (96\%)} & {\tiny (100\%)} & {\tiny (100\%)} & {\tiny (100\%)} & {\tiny (95.67\%)} \\
        \hline
        LGD & 25\% & \textbf{10\%} & 8\% & \textbf{14\%} & 73\% & 85\% & 35.83\% \\
            [-0.75ex]
             & {\tiny (99\%)} & {\tiny (84\%)} & {\tiny (97\%)} & {\tiny (100\%)} & {\tiny (100\%)} & {\tiny (100\%)} & {\tiny (96.67\%)} \\
        \hline
        LNN & 21\% & 7\% & 12\% & \textbf{14\%} & \textbf{82\%} & 87\% & \textbf{37.17\%} \\
            [-0.75ex]
             & {\tiny (99\%)} & {\tiny (81\%)} & {\tiny (100\%)} & {\tiny (100\%)} & {\tiny (100\%)} & {\tiny (100\%)} & {\tiny (95.83\%)} \\
        \hline
    \end{tabular}
    \vspace{-8mm}
    \label{tab:explanations-lunar}
\end{table}


\vspace{-2mm}
\begin{table}[h!]
    \centering
    \caption{Accuracy of competency explanations for the speed limit dataset across various true causes of low competency. Results for the pre-trained model are displayed more prominently, while results for the fine-tuned model are provided in parentheses.}
    \renewcommand{\arraystretch}{1.2}
    \begin{tabular}{|c|c|c|c|c|c|c|c|}
        \hline
        \hspace{1mm}Method\hspace{1mm} & \hspace{1mm}Spatial\hspace{1mm} & \hspace{1mm}Brightness\hspace{1mm} & \hspace{1mm}Contrast\hspace{1mm} & \hspace{1mm}Saturation\hspace{1mm} & \hspace{1mm}Noise\hspace{1mm} & \hspace{1mm}Pixelation\hspace{1mm} & \hspace{1mm}Average\hspace{1mm} \\
        \hline
        None & \textbf{2\%} & 4\% & 0\% & 0\% & 10\% & \textbf{98\%} & 19.00\% \\
            [-0.75ex]
             & {\tiny (99\%)} & {\tiny (100\%)} & {\tiny (100\%)} & {\tiny (100\%)} & {\tiny (100\%)} & {\tiny (100\%)} & {\tiny (99.83\%)} \\
        \hline
        Reco & 1\% & 12\% & \textbf{3\%} & \textbf{14\%} & \textbf{74\%} & 81\% & \textbf{30.83\%} \\
            [-0.75ex]
             & {\tiny (100\%)} & {\tiny (99\%)} & {\tiny (98\%)} & {\tiny (100\%)} & {\tiny (100\%)} & {\tiny (100\%)} & {\tiny (99.50\%)} \\
        \hline
        LGD & 0\% & \textbf{21\%} & 1\% & 12\% & 64\% & 81\% & 29.83\% \\
            [-0.75ex]
             & {\tiny (100\%)} & {\tiny (100\%)} & {\tiny (100\%)} & {\tiny (100\%)} & {\tiny (100\%)} & {\tiny (100\%)} & {\tiny (100.00\%)} \\
        \hline
        LNN & 0\% & 12\% & 1\% & 11\% & 70\% & 81\% & 29.17\% \\
            [-0.75ex]
             & {\tiny (100\%)} & {\tiny (100\%)} & {\tiny (99\%)} & {\tiny (100\%)} & {\tiny (100\%)} & {\tiny (100\%)} & {\tiny (99.83\%)} \\
        \hline
    \end{tabular}
    \vspace{-8mm}
    \label{tab:explanations-speed}
\end{table}


\subsubsection{Pre-Trained Model}

From our results using the pre-trained LLaMA model (the prominent results displayed in Tables \ref{tab:explanations-lunar} and Table \ref{tab:explanations-speed}), we observe that the explanations generated without the help of counterfactual images were only correct around one-fifth of the time. In contrast, the explanations aided by counterfactual images produced by Reco, LGD, and LNN were correct closer to one-third of the time, indicating that counterfactual images can greatly improve the accuracy of language explanations for low model competency. Examples of this improvement are provided in Figures \cref{fig:expl-saturation,fig:expl-spatial,fig:expl-contrast,fig:expl-brightness,fig:expl-noise,fig:expl-pixelation} of Appendix \ref{sec:expl_examples}. We did not observe significant differences between the Reco, LGD, and LNN methods.

It should be noted that accuracy varies substantially across the true causes of low model competency. The language model is fairly accurate at identifying noise and pixelation as causes of low competency when a counterfactual image is provided. This may be because noise and pixelation are easily observable features, and image corruption is known to reduce classification performance. The language model can also often identify anomalous objects as a cause for low model competency with the aid of a counterfactual, but the accuracy is much lower than for noise and pixelation. Although correct explanations are often generated for spatial anomalies in the lunar dataset, the language model very rarely notices digits associated with an unknown class in the speed limit dataset. The lower performance may be seen because these spatial anomalies require some high-level understanding of the training set. Finally, the language model is far more accurate in identifying brightness, contrast, and saturation as causes of low model competency when a counterfactual is provided, but accuracy still tends to be low. This poor performance may be observed because brightness, contrast, and saturation are not widely discussed causes of low model competency with which the pre-trained language model would be familiar.


While counterfactual images can greatly increase the ability to generate language explanations that correctly identify the causes of low model competency, accuracy is still not as high as we would hope, especially for particular causes of low model competency. We notice that the language model often hallucinates in its explanations--an issue commonly observed with MLLMs \cite{hallucination}. (See Figure \cref{fig:expl-hallucination} for an example of this.) We also find that the rationale for low model competency is sometimes inverted, especially when using a counterfactual, as in Figure \cref{fig:expl-inversion}.

\subsubsection{Fine-Tuned Model}

Although it may not always be practical to fine-tune the language model depending on computational constraints and availability of training data, we note that the accuracy of language explanations increases significantly after fine-tuning the model with some image-explanation pairs. (A description of the fine-tuning process is provided in Appendix \ref{sec:finetune}.) For both the lunar dataset (Table \ref{tab:explanations-lunar}) and the speed limit dataset (Table \ref{tab:explanations-speed}), we notice that the average accuracy of the fine-tuned language explanations is close to 100\% across all methods. When fine-tuning is an option, the utility of counterfactual images decreases because the model can learn reasonable explanations using only the original images. A counterfactual image may even become unhelpful for a model that has been fine-tuned well because it introduces additional variance into the data and may, in a sense, serve as a distraction to the fine-tuned language model.


%% file: secs/conclusion.tex
\section{Conclusions} \label{sec:conclusion}

In this work, we explore the use of counterfactual images to explain why an image classification model lacks confidence in its prediction. We develop five counterfactual generation methods: image gradient descent (IGD), feature gradient descent (FGD), autoencoder reconstruction (Reco), latent gradient descent (LGD), and latent nearest neighbors (LNN). We evaluate the images generated by these methods in terms of their validity, proximity, sparsity, realism, and speed across two unique datasets with six identified causes of low model competency: spatial, brightness, contrast, saturation, noise, and pixelation. While IGD and FGD generate sparse and proximal solutions, they are slow, unreliable, and tend to generate unrealistic images. Reco, LGD, and LNN tend to generate high-competency counterfactual images that appear more realistic than their original low-competency counterparts and correct for the cause of low competency observed in the original images. The best method among these three depends on the application and the properties of counterfactual images valued most highly. 


To further evaluate the utility of counterfactual images as an explanatory tool for low model competency, we develop a pipeline to generate language explanations using a pre-trained MLLM with the aid of high-competency counterfactual images. We find that, while explanations generated without the help of counterfactual images were only correct around one-fifth of the time, the explanations aided by counterfactual images produced by Reco, LGD, and LNN were correct closer to one-third of the time. This indicates that counterfactual images can greatly improve the accuracy of language explanations for low model competency. We also find that the accuracy of explanations increases to nearly 100\% after fine-tuning the MLLM with a few thousand image-explanation pairs.

%% file: secs/future.tex
\section{Limitations \& Future Work} \label{sec:future}

Although counterfactual images appear useful for explaining the reason why an image classifier lacks confidence in its prediction, much work could be done to improve the utility of these counterfactuals. Most immediately, one could more carefully select optimization parameters for the gradient descent-based methods and improve the stopping criterion. In addition, one could consider other distance metrics in the loss functions. It would also be interesting to combine these methods--in a single objective or by utilizing multiple counterfactual images. One might also design a metalearner to dynamically select the most appropriate counterfactual for an image, rather than relying on a fixed generation method.



There is also much work to be done in generating language explanations from the provided counterfactual images. First, one could evaluate other pre-trained MLLMs, beyond LLaMA. One might also explore the design of VLMs specifically for the purpose of low model competency explanation and analyze the generalizability of such methods to new datasets. It would also be beneficial to explore methods to reduce language model hallucinations--potentially through prompt engineering techniques or post-processing filters.


To more fully understand the utility of counterfactual images and language explanations, as well as how to improve them, it would be valuable to perform user studies. Future work should analyze how useful counterfactual images are to human users, allowing the user to play the role of the MLLM and evaluating how often they determine the correct cause of low competency with and without the aid of a counterfactual. It would also be interesting to receive feedback from users about the perceived utility of these counterfactuals. Similarly, users could evaluate how accurate and useful the language explanations are to them. While expanding on the analysis of counterfactual methods and explanations, it would also be useful to conduct evaluations with more diverse and complex datasets.

Finally, there remains the question of what should be done with these explanations. Going forward, it would be interesting to explore the use of counterfactual images and their language explanations as a corrective tool to improve model predictions. For example, if the model is not confident because the brightness of an image is high, perhaps the system adjusts the brightness before making a prediction. We may also use these explanations to train better models. One might use knowledge of causes of low model competency for data augmentation.

%% file: secs/appendix.tex
\newpage
\appendix

\section{Comparison of Counterfactual Images} \label{sec:counter_images}

\begin{figure}[h!]
    \centering
    \includegraphics[width=0.9\columnwidth]{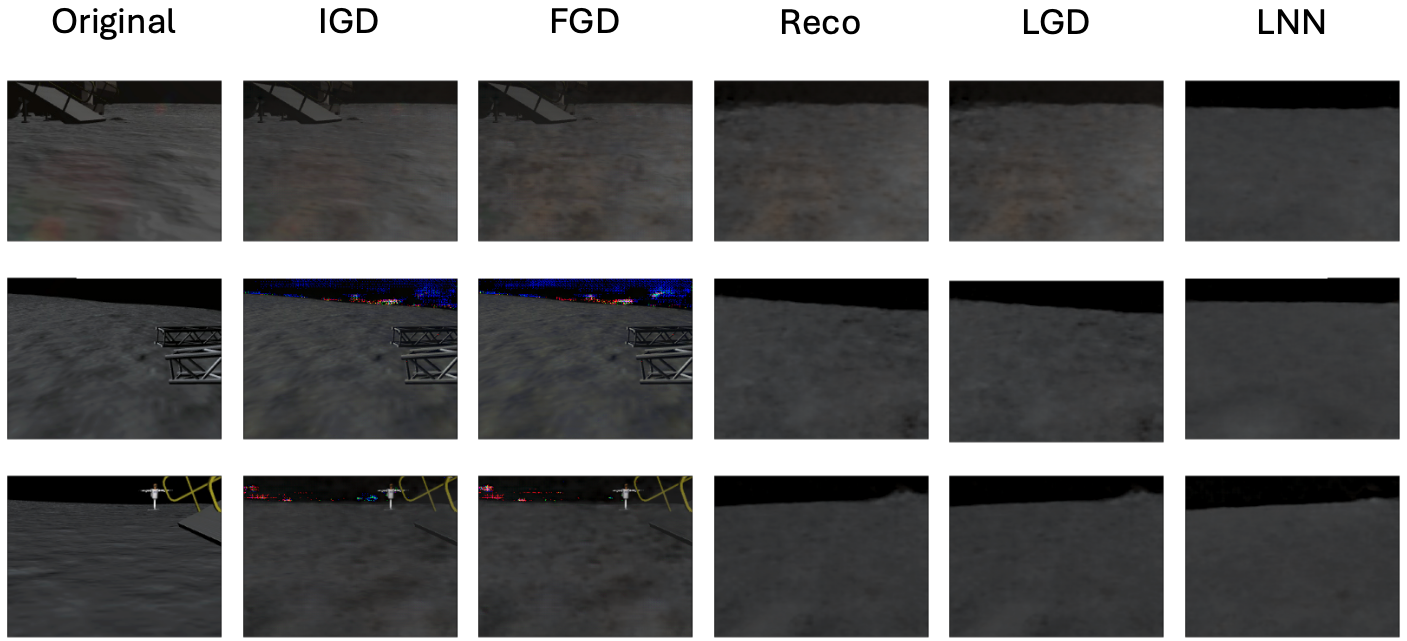}
    \caption{Counterfactual images for low-competency examples with spatial anomalies.}
    \vspace{-10mm}
    \label{fig:lunar-spatial}
\end{figure}

\begin{figure}[h!]
    \centering
    \includegraphics[width=0.9\columnwidth]{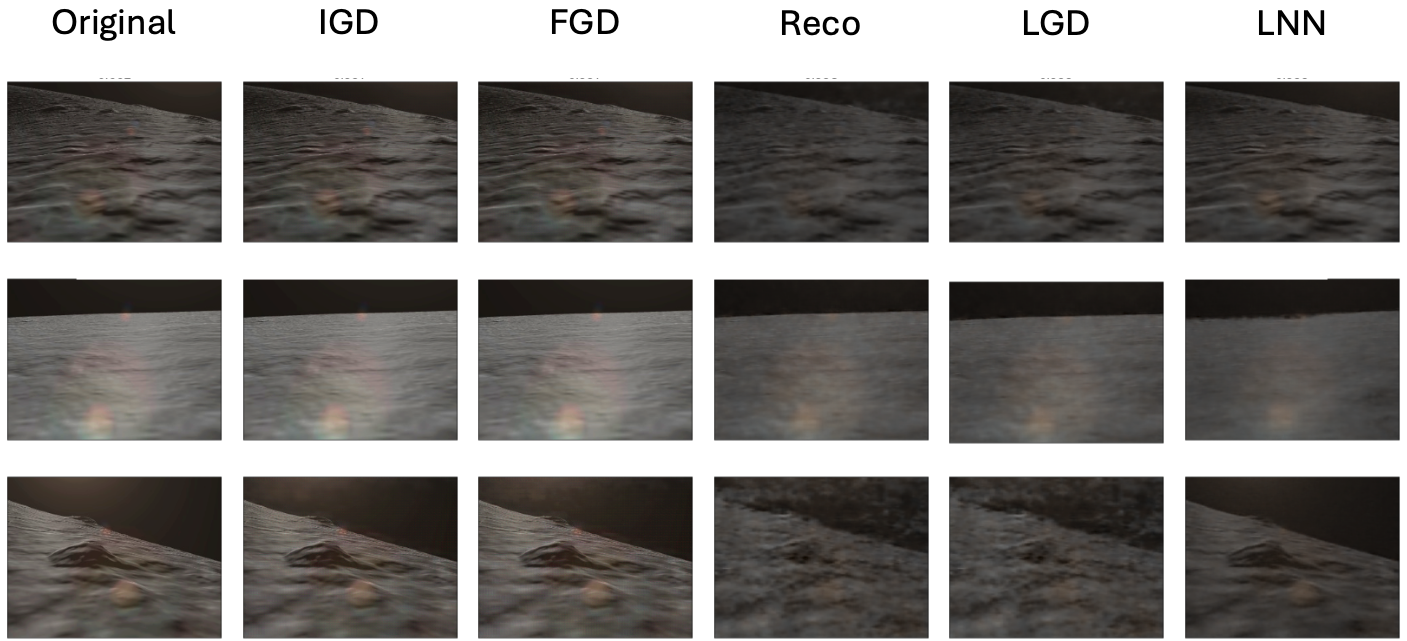}
    \caption{Counterfactual images for low-competency examples with modified brightness.}
    \vspace{-10mm}
    \label{fig:lunar-brightness}
\end{figure}

\begin{figure}[h!]
    \centering
    \includegraphics[width=0.9\columnwidth]{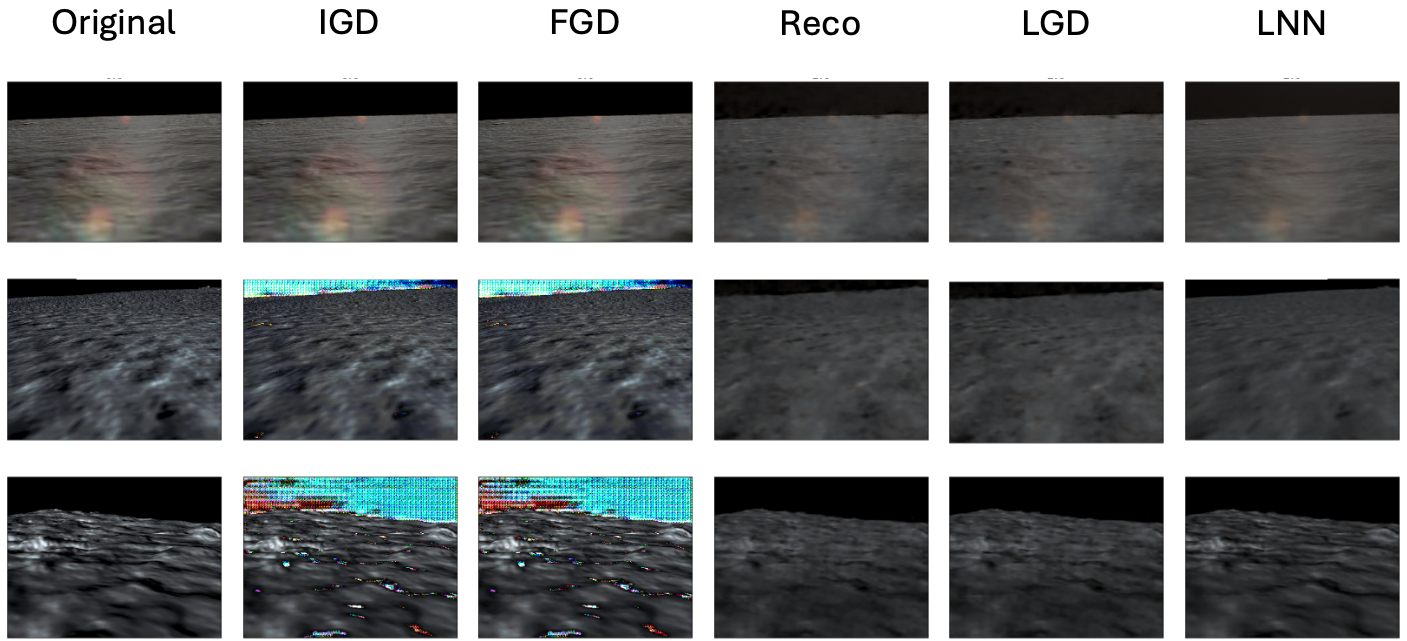}
    \caption{Counterfactual images for low-competency examples with modified contrast.}
    \vspace{-10mm}
    \label{fig:lunar-contrast}
\end{figure}

\newpage

\begin{figure}[h!]
    \centering
    \includegraphics[width=0.9\columnwidth]{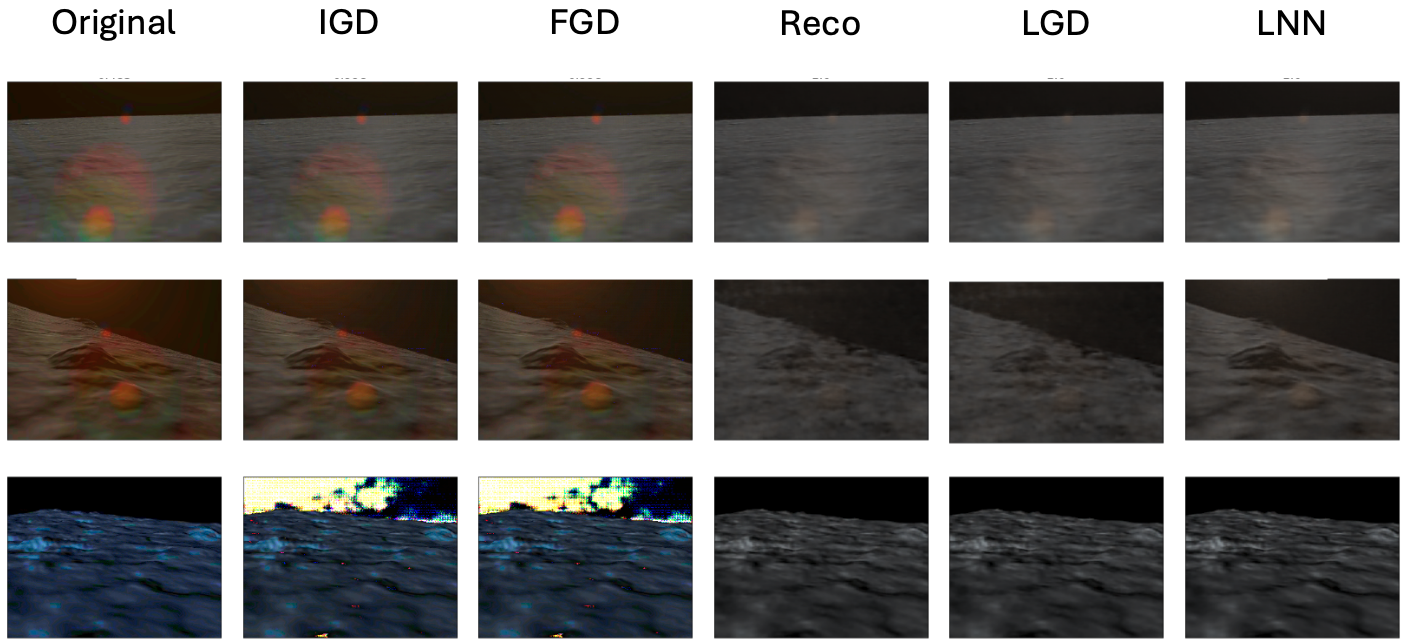}
    \caption{Counterfactual images for low-competency examples with modified saturation.}
    \vspace{-10mm}
    \label{fig:lunar-saturation}
\end{figure}

\begin{figure}[h!]
    \centering
    \includegraphics[width=0.9\columnwidth]{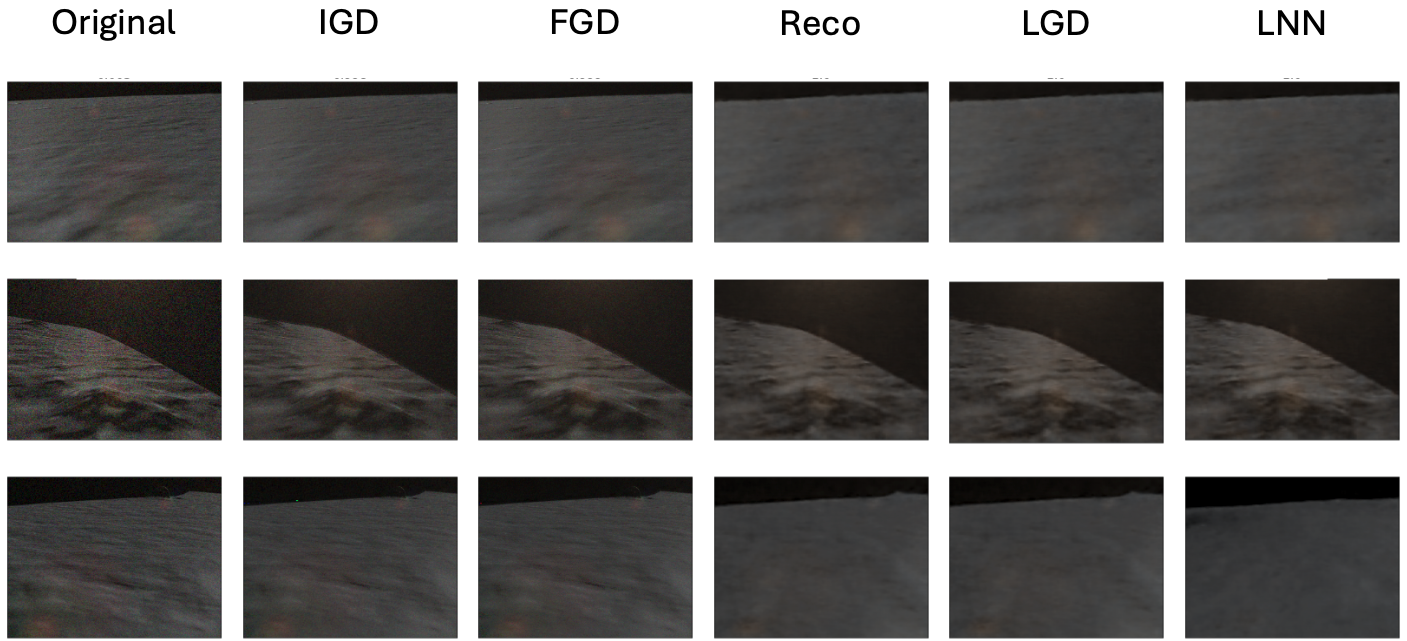}
    \caption{Counterfactual images for low-competency examples with additive noise.}
    \vspace{-10mm}
    \label{fig:lunar-noise}
\end{figure}

\begin{figure}[h!]
    \centering
    \includegraphics[width=0.9\columnwidth]{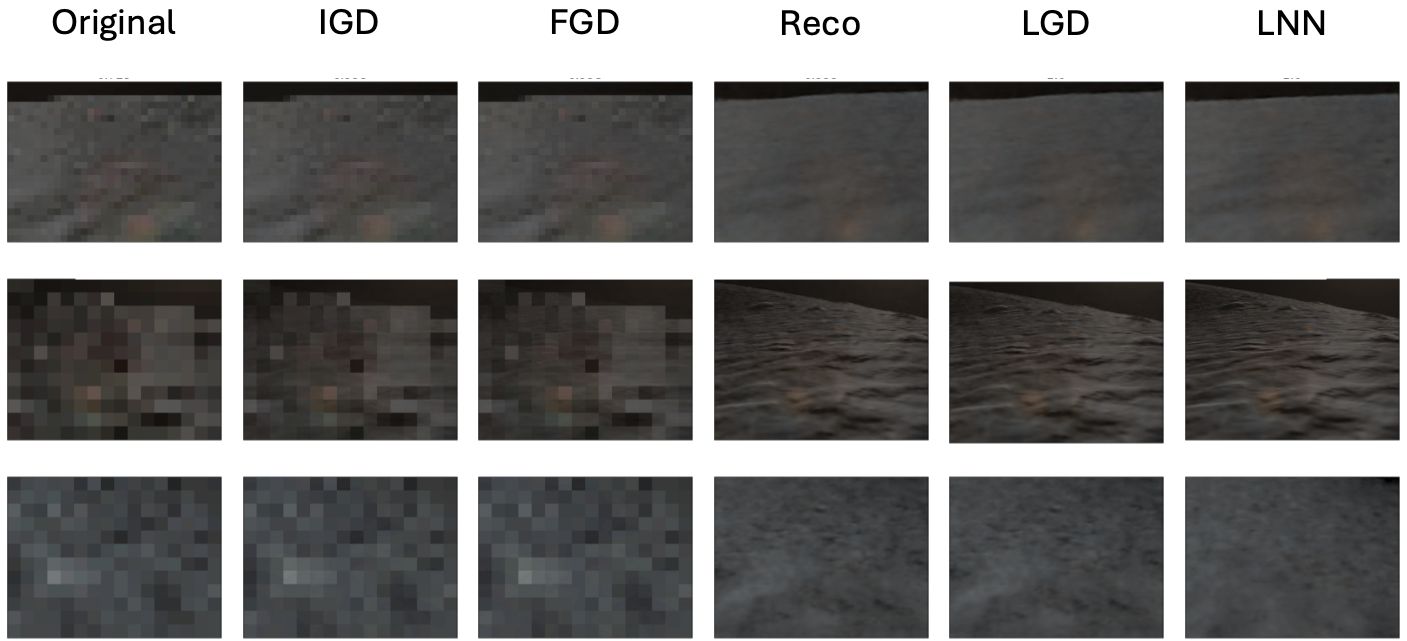}
    \caption{Counterfactual images for low-competency examples with pixelation.}
    \vspace{-10mm}
    \label{fig:lunar-pixelation}
\end{figure}

\newpage

\begin{figure}[h!]
    \centering
    \includegraphics[width=0.8\columnwidth]{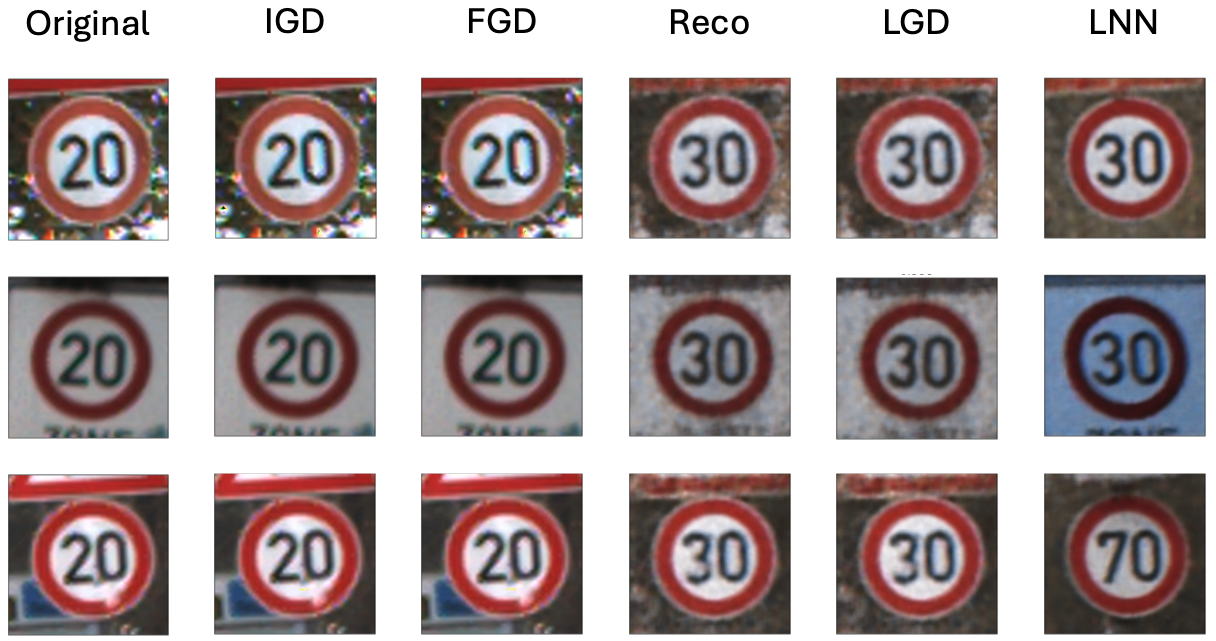}
    \caption{Counterfactual images for low-competency examples with spatial anomalies.}
    \vspace{-10mm}
    \label{fig:speed-spatial}
\end{figure}

\begin{figure}[h!]
    \centering
    \includegraphics[width=0.8\columnwidth]{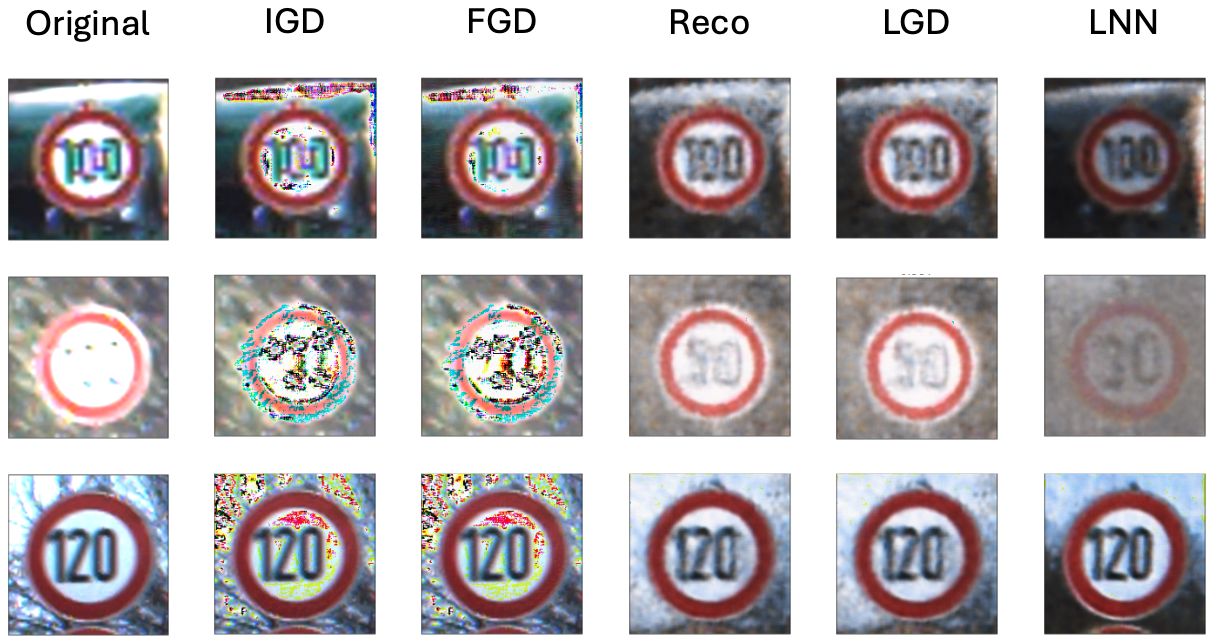}
    \caption{Counterfactual images for low-competency examples with modified brightness.}
    \vspace{-10mm}
    \label{fig:speed-brightness}
\end{figure}

\begin{figure}[h!]
    \centering
    \includegraphics[width=0.8\columnwidth]{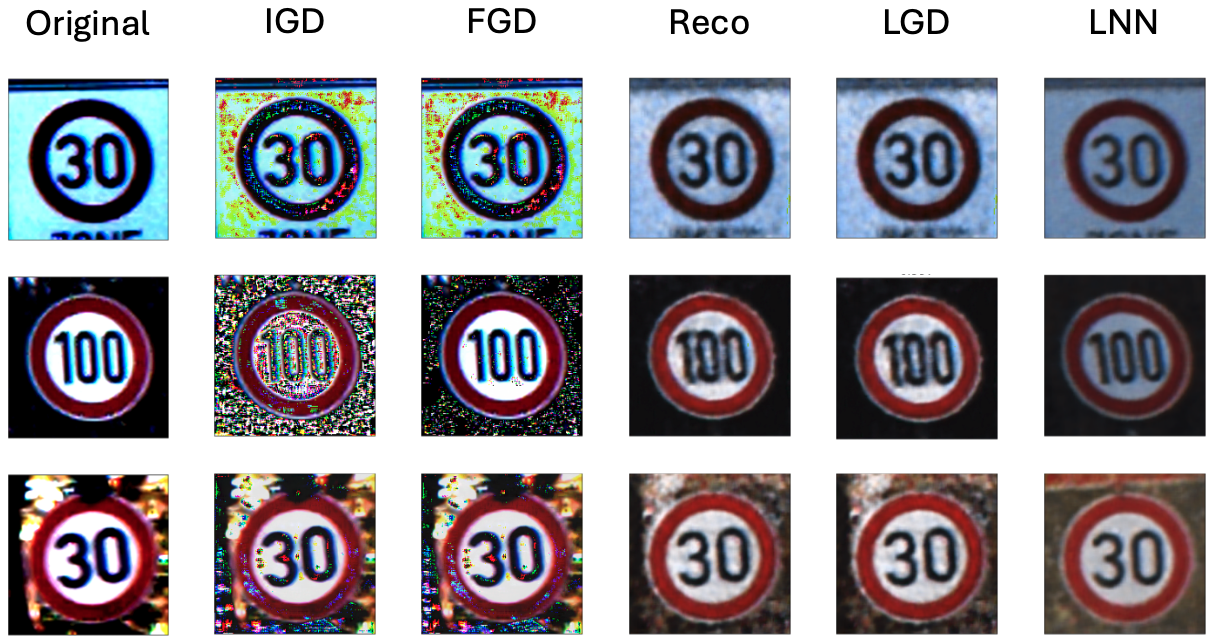}
    \caption{Counterfactual images for low-competency examples with modified contrast.}
    \vspace{-10mm}
    \label{fig:speed-contrast}
\end{figure}

\newpage

\begin{figure}[h!]
    \centering
    \includegraphics[width=0.8\columnwidth]{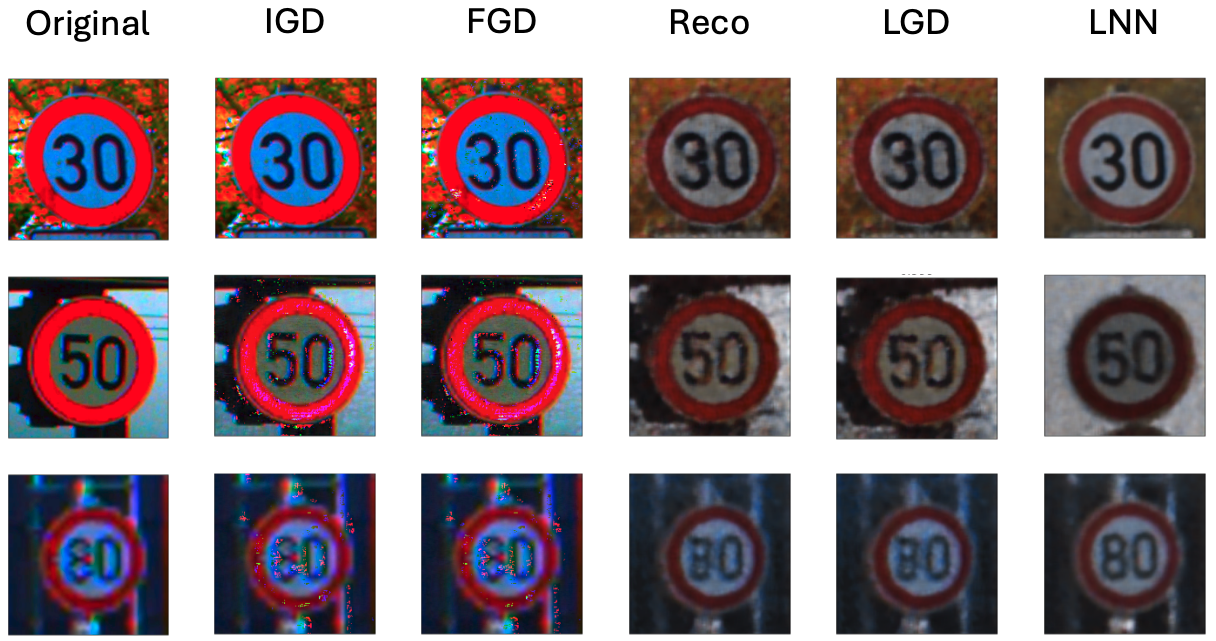}
    \caption{Counterfactual images for low-competency examples with modified saturation.}
    \vspace{-10mm}
    \label{fig:speed-saturation}
\end{figure}

\begin{figure}[h!]
    \centering
    \includegraphics[width=0.8\columnwidth]{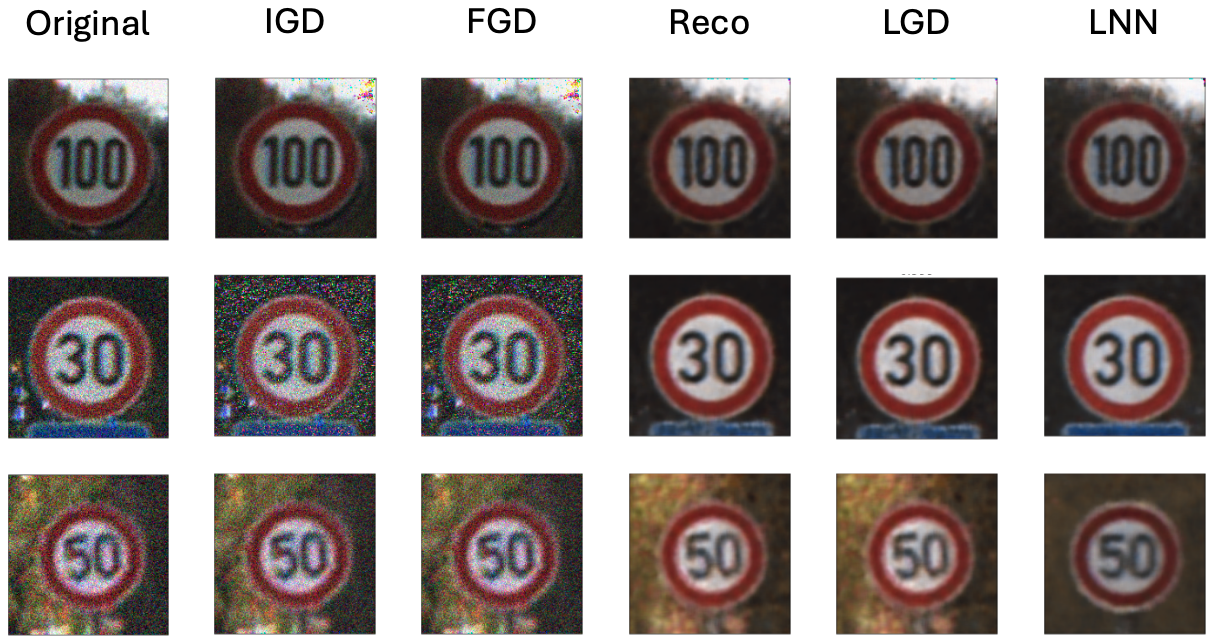}
    \caption{Counterfactual images for low-competency examples with additive noise.}
    \vspace{-10mm}
    \label{fig:speed-noise}
\end{figure}

\begin{figure}[h!]
    \centering
    \includegraphics[width=0.8\columnwidth]{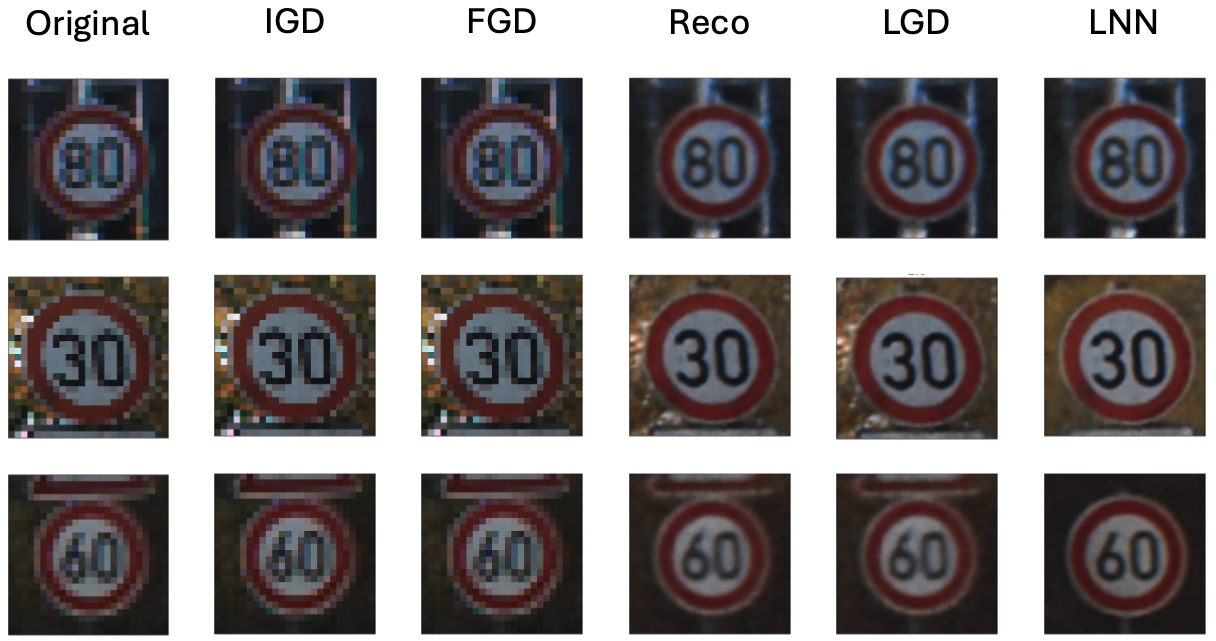}
    \caption{Counterfactual images for low-competency examples with pixelation.}
    \vspace{-10mm}
    \label{fig:speed-pixelation}
\end{figure}

\newpage

\section{Language Model Prompts} \label{sec:prompts}

\begin{promptbox}[label=prompt:dataset_lunar]{Description of Lunar Training Set}
I trained a CNN for image classification from a set of images obtained from a simulated lunar environment. The classifier learns to distinguish between different regions in this environment, such as regions with smooth terrain, regions with bumpy terrain, regions at the edge of a crater, regions inside a crater, and regions near a hill.
\end{promptbox}

\begin{promptbox}[label=prompt:dataset_speed]{Description of Speed Limit Training Set}
I trained a CNN for image classification from a dataset containing speed limit signs. The classifier learns to distinguish between seven (7) different speed limits: 30, 50, 60, 70, 80, 100, and 120 km/hr.
\end{promptbox}

\begin{promptbox}[label=prompt:competency]{Description of Competency Estimator}
In addition to the classification model, I trained a reconstruction-based competency estimator that estimates the probability that the classifier's prediction is accurate for a given image.
\end{promptbox}

\begin{promptbox}[label=prompt:instruct_none]{Instructions without Counterfactual Image}
Here is an image for which the classifier is not confident. In a single sentence, explain what properties of the image itself might lead to the observed reduction in model confidence.
\end{promptbox}

\begin{promptbox}[label=prompt:instruct_counter]{Instructions using Counterfactual Image}
Here are two images side-by-side. The first (on the left) is the original image, for which my classifier is not confident. The second image (on the right) is a similar image, for which my model is more confident. In a single sentence, explain what properties of the original image might have led to the observed reduction in model confidence.
\end{promptbox}



\section{Fine-Tuning Language Model} \label{sec:finetune}

To fine-tune the LLaMA model to generate model competency explanations for a given dataset, we first collect 3000 additional low-competency images that have not been used for training or evaluation--500 from each of the six low-competency categories. For each dataset, we gather 500 previously unseen images with spatial anomalies and generate 500 images with each of the image modifications from previously unused high-competency images. We automatically assign each of these new low-competency images a sample explanation based on their known cause of low model competency. For example, for an image with increased saturation, we assign the following explanation: ``The original image is over-saturated.''

We perform fine-tuning on a single NVIDIA GeForce RTX 4090, which has 24GB of GDDR6X RAM. To significantly reduce the size of the pre-trained model, we load the model in 4-bit quantization. To further reduce the computational effort required for fine-tuning, we use LoRA (Low-Rank Adaptation)--a technique for efficiently fine-tuning large pre-trained models by introducing learnable low-rank matrices into specific parts of the model \cite{lora}. Rather than updating the weights of the entire model, we only modify the vision layers, language layers, attention modules, and MLP modules during fine-tuning. We use a supervised fine-tuning (SFT) trainer with the Adam optimizer. We fine-tune for 10 epochs, using a training batch size of 4 and 4 gradient accumulation steps.

\section{Language Model Explanations} \label{sec:expl_examples}

\vspace{-5mm}
\begin{figure}[h!]
    \centering
    \includegraphics[width=0.9\columnwidth]{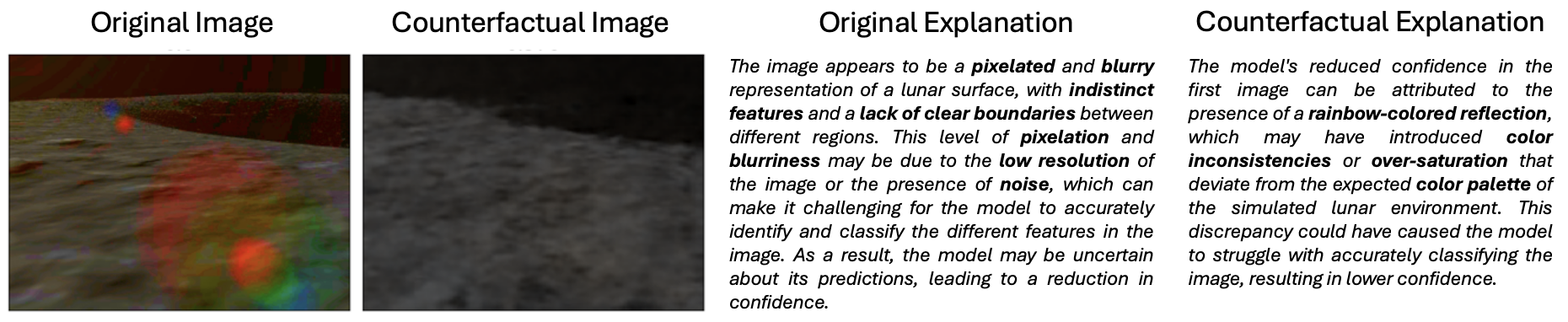}
    \caption{(i) Low-competency image with increased saturation. (ii) Counterfactual generated by Reco. (iii) Incorrect LLaMA explanation generated from original image alone. (iv) Correct LLaMA explanation generated with aid of counterfactual.}
    \vspace{-10mm}
    \label{fig:expl-saturation}
\end{figure}

\begin{figure}[h!]
    \centering
    \includegraphics[width=0.9\columnwidth]{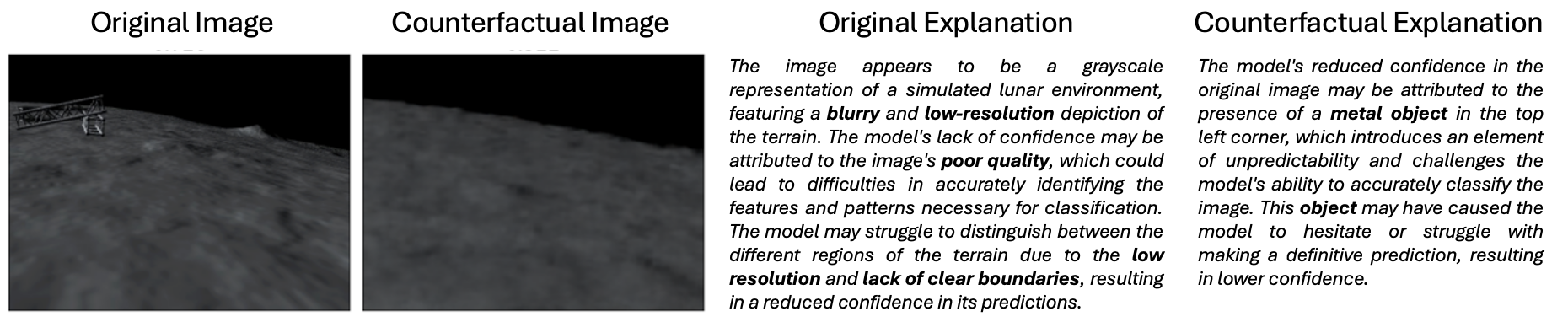}
    \caption{(i) Low-competency image with spatial anomaly (ladder at top left). (ii) Counterfactual generated by LGD. (iii) Incorrect LLaMA explanation generated from original image alone. (iv) Correct LLaMA explanation generated with aid of counterfactual.}
    \vspace{-10mm}
    \label{fig:expl-spatial}
\end{figure}

\begin{figure}[h!]
    \centering
    \includegraphics[width=0.9\columnwidth]{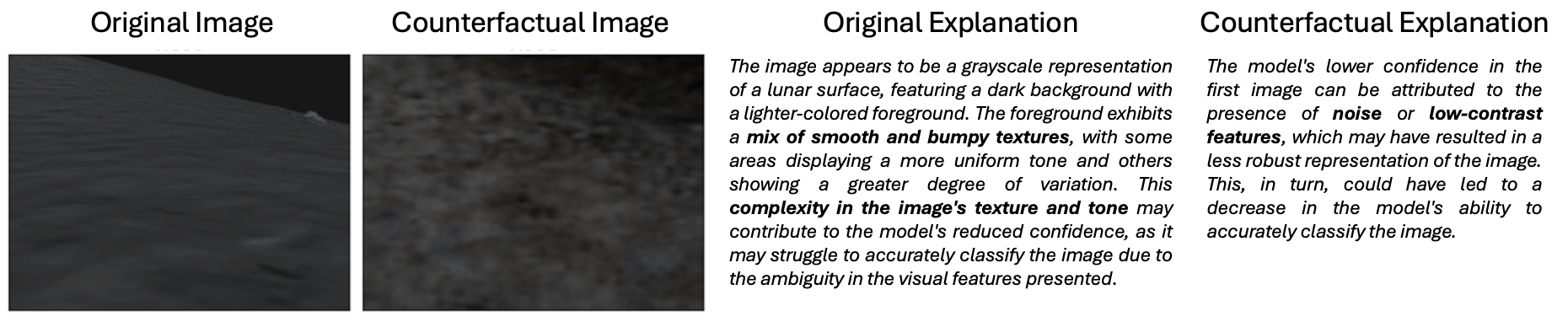}
    \caption{(i) Low-competency image with reduced contrast. (ii) Counterfactual generated by LNN. (iii) Incorrect LLaMA explanation generated from original image alone. (iv) Correct LLaMA explanation generated with aid of counterfactual.}
    \vspace{-10mm}
    \label{fig:expl-contrast}
\end{figure}

\begin{figure}[h!]
    \centering
    \includegraphics[width=0.9\columnwidth]{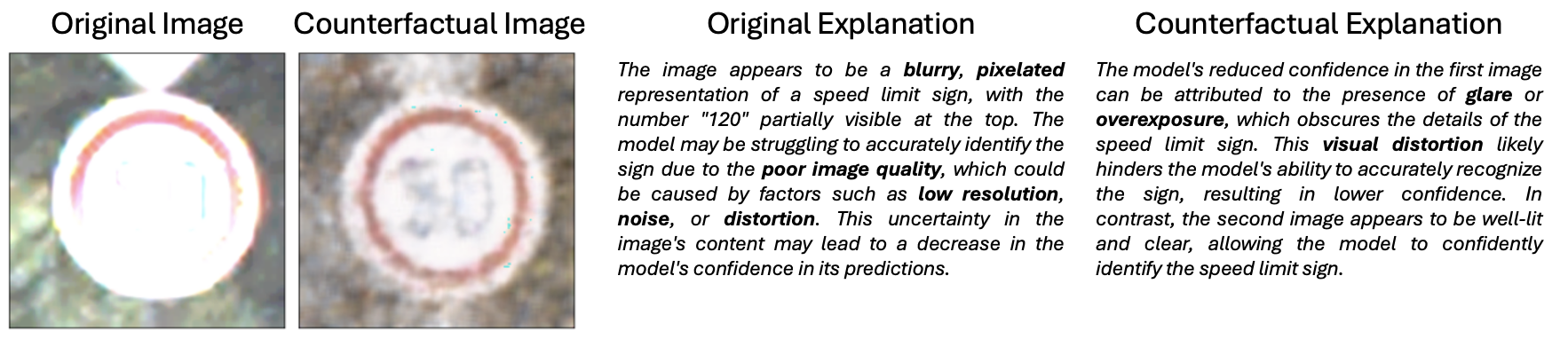}
    \caption{(i) Low-competency image with increased brightness. (ii) Counterfactual generated by Reco. (iii) Incorrect LLaMA explanation generated from original image alone. (iv) Correct LLaMA explanation generated with aid of counterfactual.}
    \vspace{-2mm}
    \label{fig:expl-brightness}
\end{figure}

\begin{figure}[h!]
    \centering
    \includegraphics[width=0.9\columnwidth]{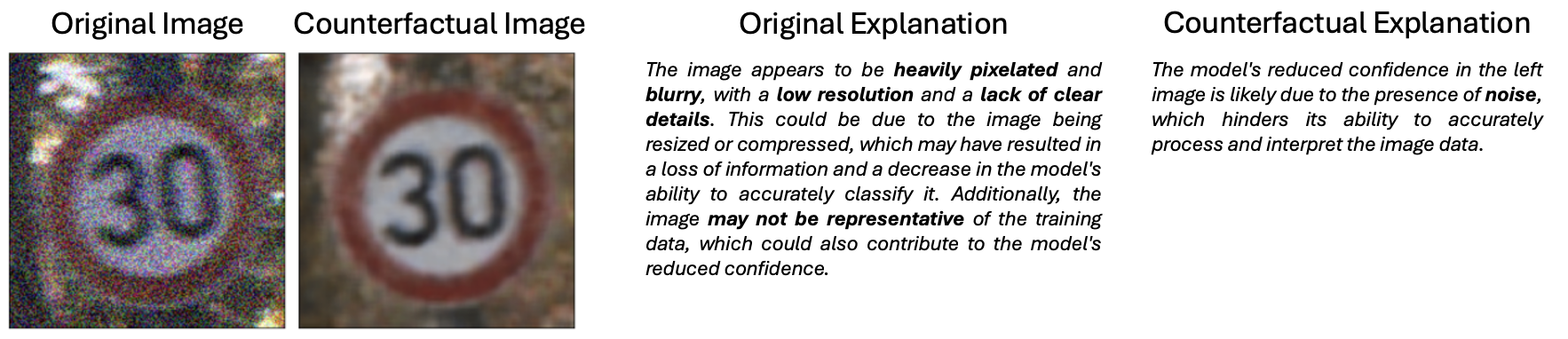}
    \caption{(i) Low-competency image with additive noise. (ii) Counterfactual generated by LGD. (iii) Incorrect LLaMA explanation generated from original image alone. (iv) Correct LLaMA explanation generated with aid of counterfactual.}
    \vspace{-2mm}
    \label{fig:expl-noise}
\end{figure}

\begin{figure}[h!]
    \centering
    \includegraphics[width=0.9\columnwidth]{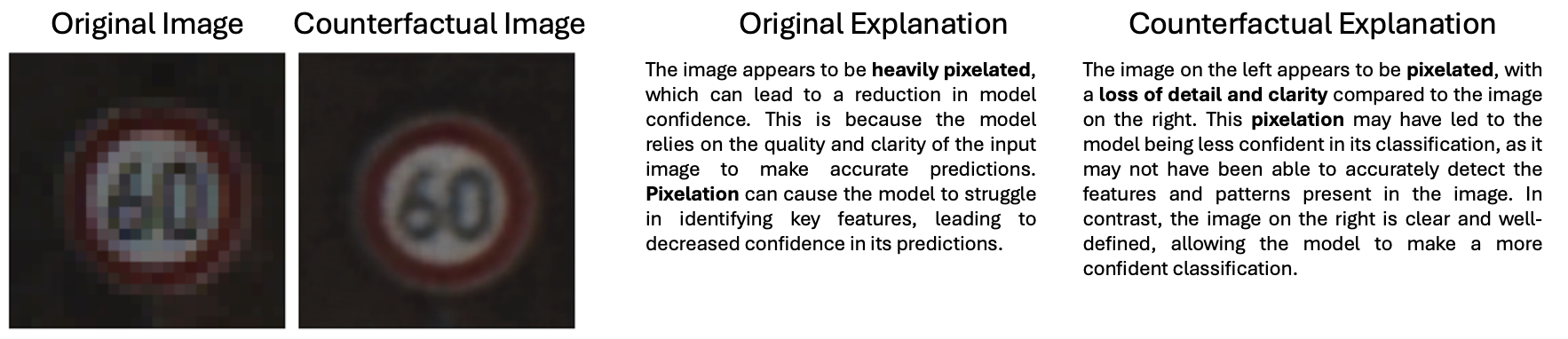}
    \caption{(i) Low-competency image with pixelation. (ii) Counterfactual generated by LNN. (iii) Correct LLaMA explanation generated from original image alone. (iv) Correct LLaMA explanation generated with aid of counterfactual.}
    \vspace{-2mm}
    \label{fig:expl-pixelation}
\end{figure}

\begin{figure}[h!]
    \centering
    \includegraphics[width=0.8\columnwidth]{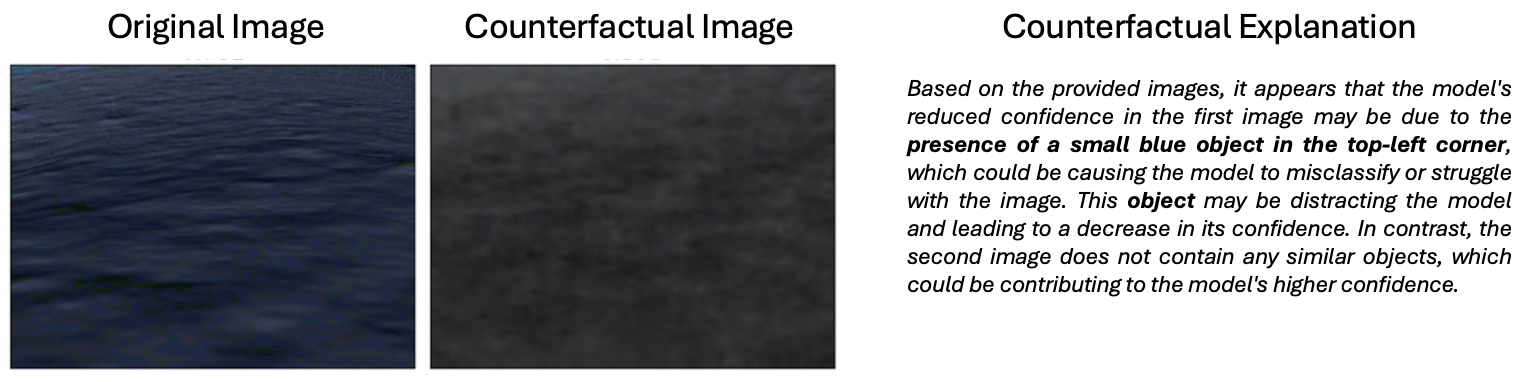}
    \caption{(i) Low-competency image with increased saturation. (ii) Counterfactual generated by LGD. (iii) Incorrect LLaMA explanation that contains a hallucination.}
    \vspace{-2mm}
    \label{fig:expl-hallucination}
\end{figure}

\begin{figure}[h!]
    \centering
    \includegraphics[width=0.8\columnwidth]{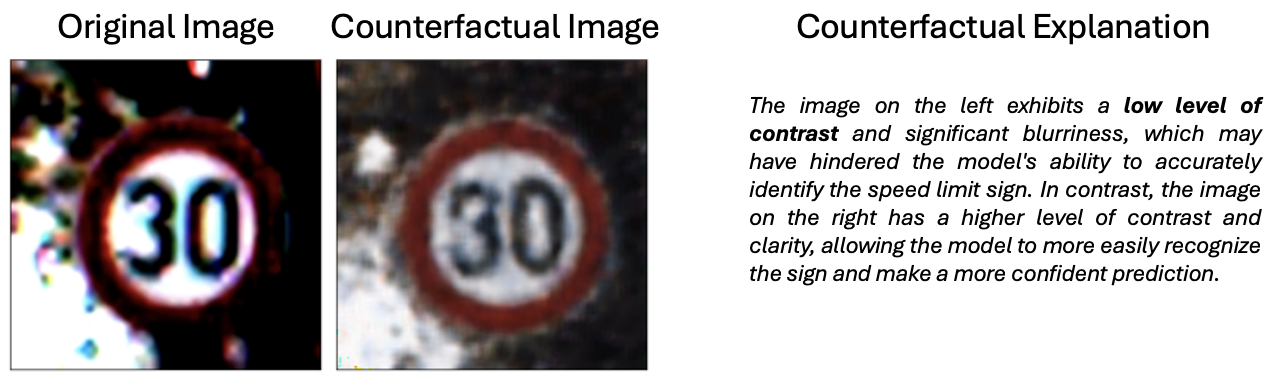}
    \caption{(i) Low-competency image with increased contrast. (ii) Counterfactual generated by Reco. (iii) LLaMA explanation that inverts the reason for low competency.}
    \vspace{-10mm}
    \label{fig:expl-inversion}
\end{figure}